\documentclass[conference]{IEEEtran}
\IEEEoverridecommandlockouts
\usepackage{cite}
\usepackage{amsmath,amssymb,amsfonts}
\usepackage{algorithmic}
\usepackage{graphicx}
\usepackage{textcomp}
\usepackage{xcolor}
\def\BibTeX{{\rm B\kern-.05em{\sc i\kern-.025em b}\kern-.08em
    T\kern-.1667em\lower.7ex\hbox{E}\kern-.125emX}}

\title{\LARGE \bf Adaptive Model Predictive Control of Wheeled Mobile Robots
\author{Nikhil Potu Surya Prakash$^{1}$ \; Tamara Perreault$^{2}$ \; Trevor Voth$^{3}$ \; and \; Zejun Zhong$^{4}$}

\thanks{The authors are graduate students in the department of Mechanical Engineering at the University of California at Berkeley, Berkeley, CA 94720, USA. Email:{\tt\small \{$^1$nikhilps,$^2$tamara\_perreault,$^3$tvoth,\\$^4$drakezhongzejun\}@berkeley.edu}}}%

\begin{document}
\maketitle
\thispagestyle{empty}
\pagestyle{empty}

\begin{abstract}
In this paper, a control algorithm for guiding a two wheeled mobile robot with unknown inertia to a desired point and orientation using an Adaptive Model Predictive Control (AMPC) framework is presented. The two wheeled mobile robot is modeled as a knife edge or a skate with nonholonomic kinematic constraints and the dynamical equations are derived using the Lagrangian approach. The inputs at every time instant are obtained from Model Predictive Control (MPC) with a set of nominal parameters which are updated using a recursive least squares algorithm. The efficacy of the algorithm is demonstrated through numerical simulations at the end of the paper.
\end{abstract}

\section{Introduction}
Model Predictive Control (MPC) as a method for designing optimal control inputs for dynamical systems has gained a lot of attention in the past few years. With the increase in computational power, using MPC to obtain optimal control inputs has become more feasible. Since MPC is a closed loop control strategy, it can robustly and optimally compensate for minor disturbances and parametric uncertainties which cannot be accounted for while computing open loop optimal control inputs.

But when the dynamic model of the system is not known or when the model is known but the physical parameters in the model are not known, the control inputs from MPC do not necessarily achieve the desired objective. It is therefore essential to first learn the model from the behaviour of the system to various inputs and then use MPC to obtain optimal control inputs. In this paper, an Adaptive Model Predictive Control (AMPC) framework for systems whose physical model is known but its parameters are unknown is presented so that the learning of parameters and obtaining the optimal control inputs is done simultaneously. More specifically, a system in which the dynamical model is linear in its physical parameters. Such a situation is very common when the robot has to picks up and carry a payload with unknown mass and mass moment of inertia like a warehouse robot which needs to pick up and transfer payloads of different masses and sizes. For such tasks, a non-adaptive control law might not do a good job and hence a necessity to learn more information about the system arises. 

Paper structure: In Section \ref{sec:Dynamics}, we discuss the nonlinear dynamics used to model the wheeled robot with nonholonomic constraints. In Section \ref{sec:AMPC}, we present the Adaptive MPC Algorithm used to control the robot along a trajectory and the parameter adaptation algorithm used to learn the parameters of the dynamical system. In Section \ref{sec:NE}, we discuss the performance of the controller as modeled by simulations.


\section{Dynamics of a wheeled mobile robot}\label{sec:Dynamics}
In this section, the dynamic evolution of the states of the wheeled mobile robot will be presented. The mechanical structure of the wheeled mobile robot is such that it is driven as a differential drive with two wheels actuated by motor torques at the rear and a smooth caster in the front keeps the robot stable. The dynamics are based on the dynamic model of a unicycle with a nonholonomic constraint that the admissible velocity of the robot is only along its orientation, thereby implying that the velocity in the direction perpendicular to its orientation is zero. The mathematical structure of the robot is isomorphic to that of a knife edge actuated by a thrust for motion control along its heading and a torque along the vector perpendicular to the motion plane of the robot to control its orientation. Using the isomorphic structure between the dynamics of the wheeled mobile robot and the knife edge, the motor torques on the wheels of the robot can be easily calculated using a one-to-one transformation. Hence to simplify the analysis and notation, a dynamic knife edge model will be used in the rest of the paper. The dynamics are described using the Lagrangian formulation.\\
The system in consideration is similar to a knife in structure but with more assumptions on the constraints that the frictional capacity available to keep the skate from moving in the direction perpendicular to its orientation.Further we assume that the frictional force along the direction of motion allows only pure rolling of the wheels and no slippage is allowed.\newline
We define two frames, an inertial frame (with subscript A) and a body fixed frame (with subscript B) in our analysis.The transformation between these two frames is achieved through the yaw angle. \\
The kinetic energy of the system with states $[x,y,\psi,\dot x,\dot y,\dot \psi]$ is given by 
\begin{equation}\label{eq:Kinetic Energy}
    T = \frac{1}{2}m \dot x^{2} + \frac{1}{2}m \dot y^2 + \frac{1}{2}J \dot \psi^2
\end{equation}
where $x$ and $y$ are the coordinates of the robot with respect to an inertial frame, $\psi$ is its heading angle and the rest are their respective derivatives.\\
We assume that the motion of the knife edge is restricted to a plane with the same gravitational potential throughout and hence there will not be any change in the potential energy of the system as the knife edge moves. But this analysis can easily be extended to a case where gravitational potential need not be constant along the path. Since the other potentials remain constant, the kinetic energy will be equal to the Lagrangian ($L:=T-V$) of the system.\\
The constraint on the motion is given by the condition that the velocity in the direction  $j_B$  $i.e.,$ along the direction perpendicular to its orientation  is $ 0 $ 
\begin{equation}\label{eq:nonholonomic constraint}
-\dot x sin\psi+ \dot y cos\psi = 0
\end{equation}
We control the system by giving two inputs on the system. A thrust $R$ in the direction of the velocity and a moment $M$ to change the orientation of the knife edge.\\
The equations of motion are derived using the Lagrangian formulation by introducing Lagrangian multipliers ($\lambda$) since the system is nonholonomic. One can even use the Appelian approach using the acceleration energy to arrive at the equations of motion. The generalized forces on the system can be derived by writing the virtual work ($\delta W$) done by the control inputs on the system.
\begin{equation}\label{eq:virtual work}
\begin{aligned}
\delta W = &(\delta x cos\psi+\delta y sin\psi)(R-bv)\\
&+\delta \psi (M-c\dot \psi)
\end{aligned}
\end{equation}
Where $v = \dot x cos\psi+\dot y sin\psi$ is the longitudinal velocity, $b$ and $c$ are the drag coefficients corresponding to the linear and angular motions respectively.
Therefore the generalized forces $Q_x$ , $Q_y$ and $Q_{\psi}$ are equal to $(R-bv)cos\psi$, $(R-bv)sin\psi$ and $M-c\dot \psi$ respectively. 
\begin{equation}\label{eq:EOMs}
\begin{aligned}
& & \frac{d}{dt} \frac{\partial{L}}{\partial{\dot x}}-\frac{\partial{L}}{\partial{x}}-\lambda a_1 = Q_x \\
& & \frac{d}{dt} \frac{\partial{L}}{\partial{\dot y}}-\frac{\partial{L}}{\partial{y}}-\lambda a_2 = Q_y \\
& & \frac{d}{dt} \frac{\partial{L}}{\partial{\dot \psi}}-\frac{\partial{L}}{\partial{\psi}}-\lambda a_3 = Q_{\psi}
\end{aligned}    
\end{equation}

The coefficients ($a_i$) to the Lagrange multipliers are the same as the coefficients of the states in the constraint equations. Substituting all the quantities and simplifying we get.
\begin{equation}\label{eq:xdd}
m\ddot x + \lambda sin\psi = (R-bv) cos\psi   
\end{equation}
\begin{equation}\label{eq:ydd}
m\ddot y - \lambda cos\psi = (R-bv) sin\psi  
\end{equation}
\begin{equation}\label{eq:M}
J\ddot \psi = M-c\dot \psi
\end{equation}
From Newton's second law, it can easily be seen that the lateral frictional force acting on the skate is the same as the Lagrangian multiplier $\lambda$. 
\begin{equation}\label{eq:R}
m\ddot x cos\psi+ m\ddot y sin \psi = R-bv
\end{equation}
\begin{equation}\label{eq:F}
-m\ddot x sin \psi + m\ddot y cos\psi = \lambda 
\end{equation}
By differentiating the constraint equation \eqref{eq:nonholonomic constraint} we get
\begin{equation}\label{eq:cdd}
-\ddot x sin\psi -\dot x \dot \psi cos\psi + \ddot y cos\psi -\dot y \dot \psi sin \psi = 0
\end{equation}
Therefore 
\begin{equation}
m\dot \psi(\dot x cos\psi+ \dot y sin\psi)  = \lambda
\end{equation}
Here $\dot x cos\psi+ \dot y sin\psi$ is the longitudinal velocity of the knife edge. Denoting the longitudinal velocity by $v$, we get
\begin{equation}
m\dot \psi v  = \lambda 
\end{equation}
Using the expression for longitudinal velocity, \eqref{eq:R} can be simplified as
\begin{equation}\label{eq:R simplified}
m\dot v = R-bv
\end{equation}
We can combine all these equations into a simple state space form as follows
\begin{equation} \label{eq:continuous dynamics}
\begin{bmatrix} 
\dot x \\
\dot y \\
\dot \psi \\
\dot v \\
\dot \omega
\end{bmatrix} = \begin{bmatrix}
v cos(\psi)\\
v sin(\psi)\\
\omega \\
\frac{(R-bv)}{m} \\
\frac{(M-c\dot \psi)}{J}
\end{bmatrix}
\end{equation}
The nonholonomic constraint that the robot cannot move in the direction perpendicular to its orientation given by \eqref{eq:nonholonomic constraint}
is satisfied at every instant and can be easily verified by substituting \eqref{eq:continuous dynamics} in \eqref{eq:nonholonomic constraint}.
To simplify the computation process by approximating the infinite dimensional optimization problem to a finite dimensional optimization problem, the dynamics will be discretized as follows and will be used to model the behavior of the robot in the rest of the paper. The discretized equations are obtained using Euler discretization with a sampling interval $\Delta t$. A more complicated yet better approximation can be obtained using first order hold or any other similar higher order interpolations for the inputs between the sampling points. The discretized dynamics are given by
\begin{equation} \label{eq:discrete dynamics}
\begin{bmatrix} 
x_{t+1} \\
y_{t+1} \\
\psi_{t+1} \\
v_{t+1}\\
\omega_{t+1}
\end{bmatrix} = \begin{bmatrix}
x_t+v_t cos(\psi_t)\Delta t\\
y_t+v_t sin(\psi_t)\Delta t\\
\psi_t+\omega_t \Delta t \\
v_t + \frac{(R-bv_t)\Delta t}{m} \\
\omega_t + \frac{(M-c\omega_t)\Delta t}{J} \\
\end{bmatrix}
\end{equation}
An equivalent representation of \eqref{eq:discrete dynamics} as following will be used in the Adaptive MPC algorithm.
\begin{equation} \label{eq:discrete dynamics params}
\begin{bmatrix} 
x_{t+1} \\
y_{t+1} \\
\psi_{t+1} \\
v_{t+1}\\
\omega_{t+1}
\end{bmatrix} = \begin{bmatrix}
x_t+v_t cos(\psi_t)\Delta t\\
y_t+v_t sin(\psi_t)\Delta t\\
\psi_t+\omega_t \Delta t \\
\alpha_v v_t + \beta_v R \\
\alpha_\omega \omega_k + \beta_\omega M \\
\end{bmatrix}
\end{equation}
Where 
\begin{equation}\label{eq:transformed system parameters}
\begin{aligned}
& \alpha_v = 1-\frac{b\Delta t}{m}\\
& \beta_v = \frac{\Delta t}{m}\\
& \alpha_\omega = 1-\frac{c\Delta t}{J}\\
& \beta_\omega = \frac{\Delta t}{J}\\
\end{aligned}
\end{equation}
It can be easily seen that by treating $\theta = [\alpha_v,\beta_v,\alpha_\omega,\beta_\omega]^T$ as proxy system parameters, the original system parameters $[m,b,J,c]^T$can be calculated using a simple one-to-one transformation. This transformation \eqref{eq:transformed system parameters} makes the system dynamics a linear function of the system parameters. This property will become handy to use existing algorithms  during the identification of the system parameters. 
These dynamics can be compactly written as
\begin{equation}\label{eq:compact dynamics}
X_{t+1} = f(X_t,u_t,\theta)
\end{equation}
where $X_k \in \mathcal{X} \subseteq \mathbb{R}^5$ is the state vector and $u_k \in \mathcal{U} \subseteq \mathbb{R}^2$ is the vector of control inputs $[R,M]^T$.The equations in \eqref{eq:discrete dynamics} will be used to describe the dynamics of the robots.

\begin{figure}[htbp]
\centerline{\includegraphics{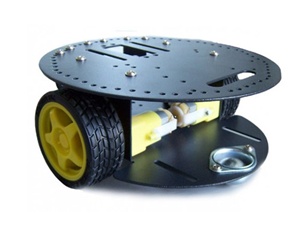}}
\caption{A sample image of a wheeled mobile robot.}
\label{WMR}
\end{figure}
\section{Adaptive MPC}\label{sec:AMPC}
In the dynamic model derived in section \ref{sec:Dynamics}, we start with no knowledge of the system parameters like the mass, mass moment of inertia and the drag coefficients of the robot and estimate them as we go along with the knowledge of the states of the system with a parameter adaptation algorithm and feed it as parameters to an optimization problem at every instance to obtain the optimal control inputs. We assume that we have full access to the states of the robot.\\
A typical MPC problem is formulated as follows

\begin{equation} \label{eq:MPC}
\begin{aligned}
U^*_t(x(t)) =\; & \underset{U_t}{\text{argmin}}
\sum_{k=0}^{N-1}q(X_{t+k},u_{t+k})    \\
\text{subject to} \;
& X_t = X(t) &(c1)\\
& X_{t+k+1} = f(X_{t+k},u_{t+k},\theta) &(c2)\\
& u_{t+k} \in \mathcal{U} &(c3)  \\
& U_t = \{u_t,u_{t+1},\dots,u_{t+N-1}\} &(c4)   \\
\end{aligned}
\end{equation}
where the constraint (c1) is the measurement, (c2) is the system model as a function of parameters, (c3) is the admissible input set and (c4) is the set of optimization variables. The control input at $t$ would be the first component of $U^*_t(x(t))$ $i.e.,$ $u^*_t$. It is worth noting that the constraints on the state are not included here as the system parameters are not known and guaranteeing the states to be confined to a desired set might not be possible. Instead one could use soft constraints for state constraints \\
MPC algorithm in \eqref{eq:MPC} is not implementable just by itself as the system parameters are unknown. So, we start with a guess of the system parameters and calculate the optimal control corresponding to the guess and update the guess at every instant using a parameter adaptation algorithm. MPC and parameter adaptation run simultaneously as shown in \ref{WMR}.The adaptive MPC structure is formulated as follows with just a minor modification where $\theta$ is replaced by $\hat \theta_t$ which is an output of the parameter adaptation algorithm.

\begin{equation} \label{eq:Adaptive MPC}
\begin{aligned}
U^*_t(x(t)) =\; & \underset{U_t}{\text{argmin}}
\sum_{k=0}^{N-1}q(X_{t+k},u_{t+k})    \\
\text{subject to} \;
& X_t = X(t) \\
& X_{t+k+1} = f(X_{t+k},u_{t+k},\hat \theta_t) \\
& u_{t+k} \in \mathcal{U}  \\
& U_t = \{u_t,u_{t+1},\dots,u_{t+N-1}\}   \\
\end{aligned}
\end{equation}

\begin{figure}[htbp]
\centerline{\includegraphics[width=95mm]{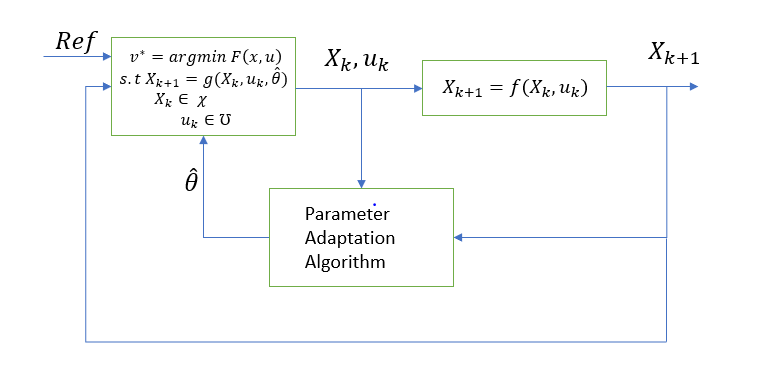}}
\caption{Adaptive MPC Algorithm}
\label{WMR}
\end{figure}

\subsection*{Parameter Adaptation Algorithm (PAA):}
A simple series parallel parameter adaptation algorithm from existing literature will be used in tandem with MPC to update the parameters, but any other PAA could be used depending on the complexity of the problem. The series parallel structure for PAA is formulated in this subsection.
Due to the transformation shown in \eqref{eq:transformed system parameters}, dynamical equations containing the system parameters can be written as an inner product of the system parameters and a regressor vector containing functions of the states.
\begin{equation}\label{regressor form of dynamics}
\begin{aligned}
& v_{t+1} = \theta_v^T\phi_{v,t} \\
& \omega_{t+1} = \theta_\omega^T\phi_{\omega,t}
\end{aligned}
\end{equation}
where $\theta_v = [\alpha_v,\beta_v]^T$ and $\theta_\omega = [\alpha_\omega,\beta_\omega]^T$ 
At every instant with the estimate of the system parameters available, an estimate of the states at the next instance can be found using
\begin{equation}\label{state estimates}
\begin{aligned}
& \hat v_{t+1} = \hat \theta_{v,t}^T\phi_{v,t} \\
& \hat \omega_{t+1} = \hat \theta_{\omega,t}^T\phi_{\omega,t}
\end{aligned}
\end{equation}
The error between the actual states measured and the projected states at the next instance can now be used to update the parameters using a series parallel PAA as 
\begin{equation}\label{PAA}
\begin{aligned}
& \hat \theta_{v,t+1} = \hat \theta_{v,t}+F_{v,t+1}\phi_{v,t}\epsilon_{v,t+1}^0 \\
& F_{v,t+1} = F_{v,t}-\frac{F_{v,t}\phi_{v,t}\phi_{v,t}^TF_{v,t}^T}{1+\phi_{v,t}^TF_{v,t}\phi_{v,t}}\\
& \hat \theta_{\omega,t+1} = \hat \theta_{\omega,t}+F_{\omega,t+1}\phi_{v,t}\epsilon_{\omega,t+1}^0 \\
& F_{\omega,t+1} = F_{\omega,t}-\frac{F_{\omega,t}\phi_{\omega,t}\phi_{\omega,t}^TF_{\omega,t}^T}{1+\phi_{\omega,t}^TF_{\omega,t}\phi_{\omega,t}}\\
\end{aligned}
\end{equation}
where $\epsilon_v^0(t+1) = v_{t+1}-\hat v_{t+1}$ and $\epsilon_\omega^0(t+1) = \omega_{t+1}-\hat \omega_{t+1}$ are the estimation errors and $\hat \theta(t) = [\hat \theta_v(t)^T \hat \theta_\omega(t)^T]^T$.
These updated system parameters at every instance can now be used in the Adaptive MPC structure in \eqref{eq:Adaptive MPC}.

\section{Numerical Experiments}\label{sec:NE}
In this section, numerical results of the algorithm developed in the previous section tested on a mobile robot with mass m = 5 kg, drag coefficient corresponding to linear motion b = 0.1 kgs$^{-1}$, Moment of inertia J = 0.2 kgm$^2$ and drag coefficient corresponding to angular motion c = 0.1 kgm$^2$s$^{-1}$ are presented. A quadratic cost with all the matrices P,Q and R as identity matrices of proper dimensions is considered. A total horizon M = 500 and a prediction horizon N = 30 is considered to evaluate the algorithm. The robot starting from the location (1,1) with a heading angle of 0 radians was asked to reach the origin with a heading of o radians.
\begin{equation} \label{eq:QMPC}
\begin{aligned}
U^*_t(x(t)) =\; & \underset{U_t}{\text{argmin}}
\frac{1}{2}e_N^{T}P_Ne_N+\sum_{k=1}^{N}\frac{1}{2}e_k^{T}Qe_k+\frac{1}{2}u_k^{T}Ru_k     \\
\text{subject to} \;
& X_t = X(t) \\
& X_{t+k+1} = f(X_{t+k},u_{t+k},\hat \theta_t) \\
& u_{t+k} \in \mathcal{U}  \\
& U_t = \{u_t,u_{t+1},\dots,u_{t+N-1}\}   \\
\end{aligned}
\end{equation}

\subsection*{Performance}
As shown in the plots in the Appendix, using the Adaptive MPC controller, the robot was able to successfully track the desired location without too much deviation. Testing different system parameters also showed that the robot was able to adapt and continue to successfully reach the goal.

%
%

\begin{figure}[htbp]
\centerline{\includegraphics[width=8cm]{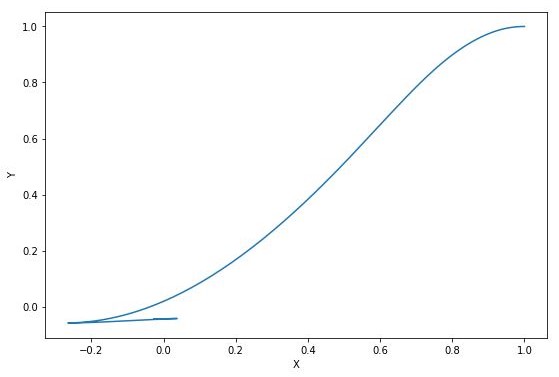}}
\caption{Final robot path (all positions are in meters)}
\label{robot_final_traj}
\end{figure}

\begin{figure}[htbp]
\centerline{\includegraphics[width=8cm]{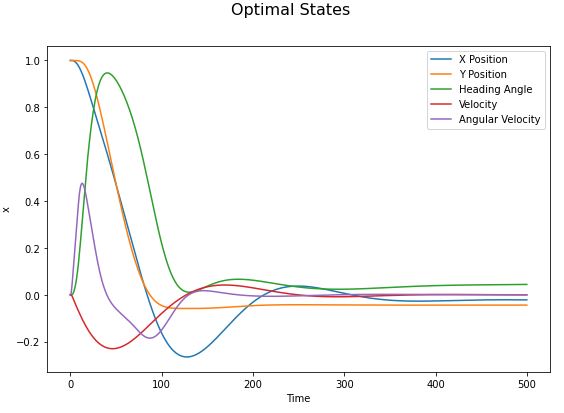}}
\caption{Optimal states of the robot with Adaptive MPC (All the states are in SI units and the time is in deci seconds to match the iteration number)}
\label{opt_states_adapt}
\end{figure}

\begin{figure}[htbp]
\centerline{\includegraphics[width=8cm]{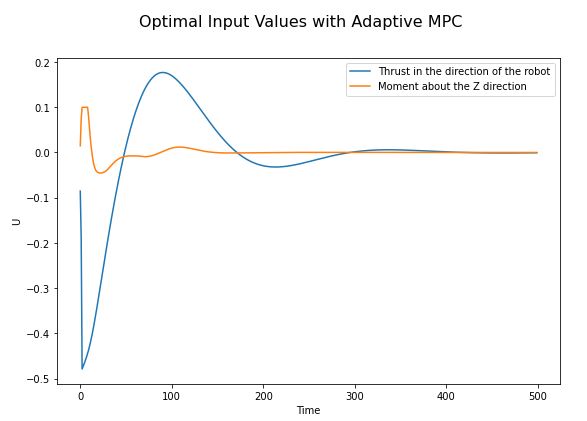}}
\caption{Optimal input values with no knowledge of the system parameters (Both the inputs are in SI units)}
\label{input_adapt}
\end{figure}

\begin{figure}[htbp]
\centerline{\includegraphics[width=8cm]{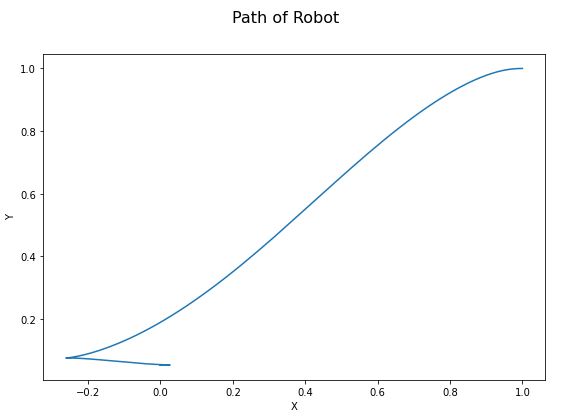}}
\caption{Optimal path of the robot with MPC with full knowledge of the system}
\label{robot_traj_noadapt}
\end{figure}

\begin{figure}[htbp]
\centerline{\includegraphics[width=8cm]{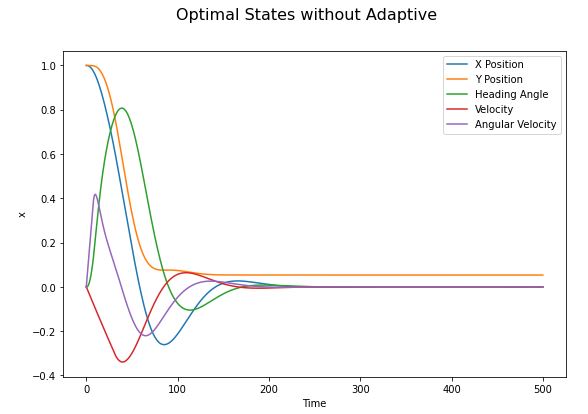}}
\caption{Optimal states of the robot with MPC with full knowledge of the system}
\label{opt_states_noadapt}
\end{figure}

\begin{figure}[htbp]
\centerline{\includegraphics[width=8cm]{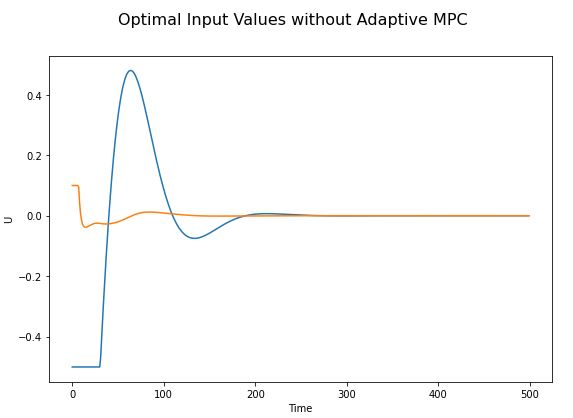}}
\caption{Optimal inputs of the robot with MPC with full knowledge of the system}
\label{opt_input_noadapt}
\end{figure}

\section*{Conclusion}
In this work, adaptive MPC was shown to be an effective strategy for optimal control of a simple wheeled robot with unknown parameters. Utilizing a nonlinear system model and introducing an adaptive element to the MPC to control for unknown system parameters makes it more difficult to show persistent feasibility. However, our simulation results look reasonably accurate and have very interesting applications. The goal tracking problem can easily be extended for trajectory tracking by modifying the cost function to minimize the distance from the trajectory and soft constraints can also be introduced on the states of the system.









\begin{thebibliography}{99}
\bibitem{c1} Francesco Borrelli, Alberto Bemporad, Manfred Morari, "Predictive control for linear and hybrid systems", Cambridge University Press.
\bibitem{c2} Jason Kong, Mark Pfeiffer, Georg Schildbach, Francesco Borrelli, "Kinematic and dynamic vehicle models for autonomous driving control design", in IEEE Intelligent Vehicles Symposium (IV), 2015, pp 1094--1099.
\bibitem{c3} Veronica Adetola, Darryl DeHaan, Martin Guay,
"Adaptive model predictive control for constrained nonlinear systems", in
Systems and Control Letters,
Volume 58, Issue 5,
2009,
Pages 320--326.
\bibitem{c4} Hiroaki Fukushima, Tae-Hyoung Kim, Toshiharu Sugie,"Adaptive model predictive control for a class of constrained linear systems based on the comparison model", in Automatica,
Volume 43, Issue 2,
2007,
Pages 301--308.
\bibitem{c5} Tae-Hyoung Kim, H. Fukushima and T. Sugie, "Robust adaptive model predictive control based on comparison model," 2004 43rd IEEE Conference on Decision and Control (CDC) (IEEE Cat. No.04CH37601), 2004, pp. 2041-2046 Vol.2, doi: 10.1109/CDC.2004.1430348.
\bibitem{c6} K. Pereida and A. P. Schoellig, "Adaptive Model Predictive Control for High-Accuracy Trajectory Tracking in Changing Conditions," 2018 IEEE/RSJ International Conference on Intelligent Robots and Systems (IROS), 2018, pp. 7831-7837, doi: 10.1109/IROS.2018.8594267.
\bibitem{c7} Anthony M Bloch, "Nonholonomic mechanics and control", Springer, New York, NY, 2003.
\bibitem{c8} AM Bloch, PS Krishnaprasad, JE Marsden, RM Murray, "Nonholonomic mechanical systems with symmetry", in Archive for Rational Mechanics and Analysis, Vol 136, Issue 1, 1996, pp 21--99.
\bibitem{c9} Lennart Ljung, Keith Glover, "Frequency domain versus time domain methods in system identification", in Automatica, Vol. 17, Issue 1, 1981, pp. 71-86.
\bibitem{c10} MATLAB. version 9.10.0 (R2016b). Natick, Massachusetts: The MathWorks Inc.,2016.
\bibitem{c11} Zhang, Qinghua. “Using Wavelet Network in Nonparametric Estimation”, IEEE Transactions on Neural Networks 8, no. 2 , March 1997. https://doi.org/10.1109/72.557660.
\bibitem{c12} Sjöberg, Jonas, Qinghua Zhang, Lennart Ljung, Albert Benveniste, Bernard Delyon, Pierre-Yves Glorennec, Håkan Hjalmarsson, and Anatoli Juditsky. “Nonlinear Black-Box Modeling in System Identification: A Unified Overview”, Automatica 31, no. 12 (December 1995): 1691–-1724. https://doi.org/10.1016/0005-1098(95)00120-8.
\bibitem{c13} Ljung, Lennart, and Torkel Glad. Modeling of Dynamic Systems. Prentice Hall Information and System Sciences Series. Englewood Cliffs, NJ: PTR Prentice Hall, 1994.
\bibitem{c14} Ljung, Lennart. System Identification: Theory for the User. Second edition. Prentice Hall Information and System Sciences Series. Upper Saddle River, NJ: PTR Prentice Hall, 1999.
\bibitem{c15} Söderström, Torsten, and Petre Stoica. System Identification. Prentice Hall International Series in Systems and Control Engineering. New York: Prentice Hall, 1989.
\bibitem{c16} Lennart Ljung, "Analysis of recursive stochastic algorithms", IEEE transactions
on automatic control 22.4, 1977, pp. 551--575.

\end{thebibliography}
\end{document}